\PassOptionsToPackage{unicode}{hyperref}
\PassOptionsToPackage{hyphens}{url}
\documentclass[
]{article}
\usepackage{xcolor}
\usepackage{amsmath,amssymb}
\usepackage[T1]{fontenc}
\usepackage[utf8]{inputenc}
\usepackage{textcomp}
\usepackage{lmodern}
\IfFileExists{upquote.sty}{\usepackage{upquote}}{}
\IfFileExists{microtype.sty}{
  \usepackage[]{microtype}
  \UseMicrotypeSet[protrusion]{basicmath} 
}{}
\makeatletter
\@ifundefined{KOMAClassName}{
  \IfFileExists{parskip.sty}{%
    \usepackage{parskip}
  }{
    \setlength{\parindent}{0pt}
    \setlength{\parskip}{6pt plus 2pt minus 1pt}}
}{
  \KOMAoptions{parskip=half}}
\makeatother
\usepackage{longtable,booktabs,array}
\usepackage{multirow}
\usepackage{wrapfig}
\usepackage{changepage}
\usepackage{adjustbox}
\usepackage{makecell}
\usepackage{titling}
\usepackage{times}
\usepackage[top=1in, bottom=1in, left=1.5in, right=1.5in]{geometry}
\usepackage{calc} 
\usepackage{etoolbox}
\makeatletter
\patchcmd\longtable{\par}{\if@noskipsec\mbox{}\fi\par}{}{}
\makeatother
\IfFileExists{footnotehyper.sty}{\usepackage{footnotehyper}}{\usepackage{footnote}}
\makesavenoteenv{longtable}
\usepackage{graphicx}
\makeatletter
\newsavebox\pandoc@box
\newcommand*\pandocbounded[1]{
  \sbox\pandoc@box{#1}%
  \Gscale@div\@tempa{\textheight}{\dimexpr\ht\pandoc@box+\dp\pandoc@box\relax}%
  \Gscale@div\@tempb{\linewidth}{\wd\pandoc@box}%
  \ifdim\@tempb\p@<\@tempa\p@\let\@tempa\@tempb\fi
  \ifdim\@tempa\p@<\p@\scalebox{\@tempa}{\usebox\pandoc@box}%
  \else\usebox{\pandoc@box}%
  \fi%
}
\def\fps@figure{htbp}
\makeatother
\usepackage{bookmark}
\IfFileExists{xurl.sty}{\usepackage{xurl}}{} 
\usepackage{hyperref}
\hypersetup{
  hidelinks,
  pdfcreator={LaTeX via pandoc}}

\newcommand{\papertitle}{\textbf{The Data Efficiency Frontier of Financial Foundation Models: Scaling Laws from Continued Pretraining}}
\newcommand{\paperauthors}{\textbf{Jesse Ponnock}}
\newcommand{\paperaffiliations}{
Johns Hopkins University\\
\texttt{jponnoc1@jh.edu}\\
}

\usepackage{placeins}
\usepackage{caption}

\begin{document}

\begin{center}
\rule{\textwidth}{4pt} \\[1.25em]  
\LARGE \textbf{\papertitle} \\[0.25em]
\rule{\textwidth}{1pt} \\[1.25em]
\large \textbf{\paperauthors} \\[0.5em]
\small \paperaffiliations
\end{center}

\vspace{1.5em}

\begin{center}
\Large \textbf{Abstract}
\end{center}

\vspace{.75em}

\begin{adjustwidth}{3.5em}{3.5em}
Domain-adaptive pretraining (DAPT) offers a practical path to specializing large language models for high-value domains without full retraining. We conduct an early-stage scaling-law analysis of continued pretraining on U.S. SEC filings, training 1B and 3B-parameter Llama-3.2 models on a 400M-token financial corpus with validation checkpoints at 50M, 100M, 200M, and 400M tokens. Results show consistent improvements in SEC-domain validation loss for both models, with the largest gains occurring within the first 200M tokens and diminishing returns thereafter. Power-law fits reveal shallow exponents, indicating that financial language is highly regular and efficiently learnable under continued pretraining. General-domain validation loss remains effectively unchanged across all token budgets, suggesting minimal drift and no signs of catastrophic forgetting. A data-efficiency frontier further shows that both models move toward improved specialization with negligible mixed-domain degradation. Together, these findings provide early empirical guidance for scaling financial foundation models, suggesting that meaningful domain adaptation can be achieved with comparatively modest token budgets and that larger model scales (7B–70B) remain tractable under projected data requirements.
\end{adjustwidth}

\section{Introduction}\label{sec:introduction}

DAPT provides a mechanism for improving model performance in specialized domains by exposing an existing pretrained model to additional text from a narrow distribution. In this study, we treat DAPT on SEC filings as an early-stage scaling-law experiment. To accomplish this, we train with a total token budget of 400M tokens. During training, validation loss is logged at key points, providing effective dataset sizes of 50M, 100M, 200M, and 400M tokens. This enables the examination of how loss decreases as a function of additional data, similar in spirit to the token-scaling methodology used in Chinchilla \cite{hoffmann2022chinchilla} and related studies.

Our goal is to understand whether continued pretraining on SEC filings yields measurable domain improvements, how quickly the marginal benefit of additional data diminishes, and whether specialization induces degradation on mixed-domain text. To this end, we evaluate two model sizes—Llama-3.2-1B and Llama-3.2-3B—using domain validation loss, mixed-domain validation loss, power-law fits of loss as a function of tokens, and a data-efficiency frontier characterizing the tradeoff between specialization and forgetting.

More broadly, this analysis provides insight into the data efficiency of financial foundation models. Understanding how much domain-specific data is required for meaningful adaptation and where diminishing returns begin helps clarify how such models should be scaled, specialized, and deployed in real world economic applications.

\section{Background}\label{sec:background}
Decoder-only transformers learn to predict the next token in a sequence, and their performance is tied closely to the size and diversity of their pretraining corpus. When a pretrained model is exposed to additional domain-specific data, it can shift its internal representation to better model that distribution, a process widely referred to as DAPT.

DAPT has been shown to improve performance in biomedical \cite{lee2020biobert}, legal \cite{chalkidis2020legalbert}, and safety regulatory domains \cite{yang2025msglm}. While several financial-domain models exist, such as FinBERT \cite{yang2020finbert} and BloombergGPT \cite{wu2023bloomberggpt}, these are either encoder-only architectures (FinBERT) or trained from scratch rather than through continued pretraining (BloombergGPT). In contrast, there is relatively little systematic analysis of how modern decoder-only LLMs respond to domain-adaptive continued pretraining, particularly in financial text settings.

Two practical issues motivate this study. First, we examine whether continued pretraining on SEC filings meaningfully reduces domain-specific loss, given the structured nature of financial disclosures. Second, we evaluate whether such specialization degrades performance on mixed-domain text, a form of catastrophic forgetting relevant for practical deployments.

Scaling law analysis also helps interpret the magnitude of improvements. In linear space, loss curves from continued pretraining can appear visually flat, even when the model is still learning. Plotting loss against tokens in log-log space reveals the underlying power law relationship, making small but consistent gains more apparent and indicating whether additional data is likely to yield further improvement.

This study examines all three perspectives—domain performance, general domain stability, and scaling-law behavior—to characterize DAPT effectiveness on SEC filings.

\section{Methods}\label{sec:methods}

\subsection{Dataset}\label{sec:dataset}

We constructed a 400M token corpus from three SEC filing types—10-K, 10-Q, and DEF 14A (proxy statements). To ensure broad sectoral coverage, we began with the SEC’s Company Ticker–CIK mapping dataset \cite{sec2024tickers} and selected the first 1,000 companies listed. For each company, we collected ten years of filings directly from EDGAR.

Raw filings were downloaded and converted to text form using EdgarTools \cite{demiroglu2024edgartools} and then passed through a multi-stage preprocessing pipeline. Given the heavy use of tables, formatting artifacts, and inline XBRL tagging in SEC disclosures, we used a conservative whitelist approach for text extraction. Only long form narrative content was retained, including Management’s Discussion \& Analysis (MD\&A), Risk Factors, Business Overview, Compensation Discussion \& Analysis, and Notes to the Financial Statements. Content that could not be reliably converted into clean prose was discarded. Approximately 9.6\% of downloaded filings were excluded at this stage because the cleaning pipeline yielded insufficient extractable text.

All remaining long form text was aggregated and deduplicated. Because SEC filings contain substantial repetition across years and issuers, we applied MinHash locality sensitive hashing (LSH) \cite{broder1997minhash} for approximate near-duplicate detection. MinHash LSH hashes text segments into signatures such that documents with high Jaccard similarity are mapped to the same buckets with high probability. Segments flagged as near-duplicates were removed. This procedure eliminated approximately 1.9\% of tokens.

The resulting cleaned corpus was tokenized using the native Llama 3 tokenizer. Text was packed into sequences of 1024 tokens, and the process was halted upon reaching a total of 400 million tokens, forming the final dataset used for continued pretraining.

\subsection{Models}\label{sec:models}

We evaluated continued pretraining on two decoder-only models: Llama-3.2-1B and Llama-3.2-3B. In both cases, all parameters were updated during training, with no adapters, frozen layers, or parameter-efficient tuning mechanisms. Using two adjacent model sizes allowed a controlled comparison of parameter scaling effects under identical data budgets and training conditions.

\subsection{Training Setup}\label{sec:training-setup}

Models were trained for a single epoch over a 400M token corpus using a 1024 token context length, the AdamW optimizer, and a fixed $5\times10^{-6}$ learning rate with no scheduler. Training used a batch size of 8, and validation loss was recorded approximately every 25 million tokens. All experiments were run on a single NVIDIA H100 GPU, with the 1B model completing in 9 hours 27 minutes and the 3B model completing in 9 hours 52 minutes.

\subsection{Evaluation Metrics}\label{sec:evaluation-merics}

During training, the model logged three loss signals:
\begin{itemize}
    \item \textbf{SEC domain validation loss}, computed on a fixed 4M-token validation set derived from SEC filings.
    \item \textbf{General domain validation loss}, computed on a separate 4M-token validation set constructed from Wikipedia text to measure out-of-domain stability.
    \item \textbf{Training loss}, recorded at each validation interval to track learning dynamics.
\end{itemize}

These losses allow us to assess domain improvement, general domain drift, and overall training behavior.

\section{Results} \label{sec:results}

We analyze model behavior across four approximate token budgets of 50M, 100M, 200M, and 400M tokens, corresponding to the logged validation checkpoints during continued pretraining. Although the full learning curves are plotted continuously, our interpretation follows standard scaling law methodology by focusing on these discrete training milestones. This allows direct comparison of loss improvements as additional data is consumed.

\subsection{SEC Domain Validation Loss}\label{sec:sec-domain-validation-loss}

\vspace{0.6em}

\begin{center}
\begin{minipage}{0.9\linewidth}
    \centering
    \includegraphics[width=0.75\linewidth]{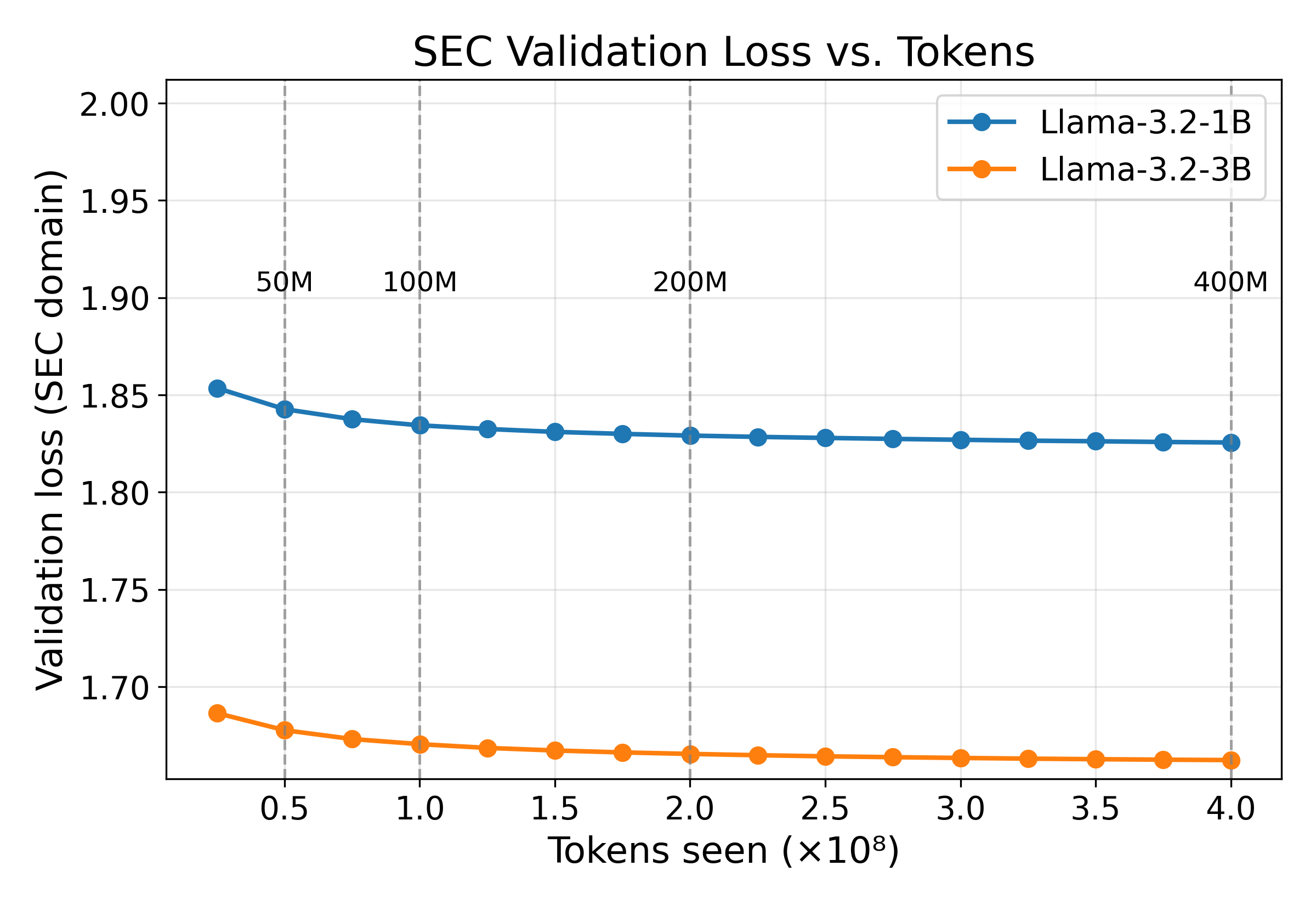}
    \vspace{-0.6em}
    \captionof{figure}{SEC-domain validation loss for 1B and 3B models}
    \label{fig:sec-loss}
\end{minipage}
\end{center}

\vspace{0.75em}

SEC domain validation loss decreases smoothly as additional SEC tokens are consumed. Both models exhibit their largest improvements in the early stages of continued pretraining: loss drops noticeably between 50M and 200M tokens, after which the rate of improvement slows. Beyond roughly 250M tokens, both curves begin to flatten, suggesting diminishing marginal returns under the current training setup.

Across all token budgets, the 3B model consistently outperforms the 1B model. It starts with a lower initial loss and maintains a stable gap at every milestone, reflecting the expected capacity advantage documented in prior scaling law studies. The overall pattern of early gains followed by a tapering curve mirrors classical token scaling behavior and indicates that SEC-focused DAPT continues to yield benefit, but with decreasing efficiency as the model saturates the domain signals available in the corpus.

\subsection{Scaling-Law Behavior}\label{sec:scaling-law-behavior}

\vspace{0.6em}

\begin{center}
\begin{minipage}{0.9\linewidth}
    \centering
    \includegraphics[width=0.82\linewidth]{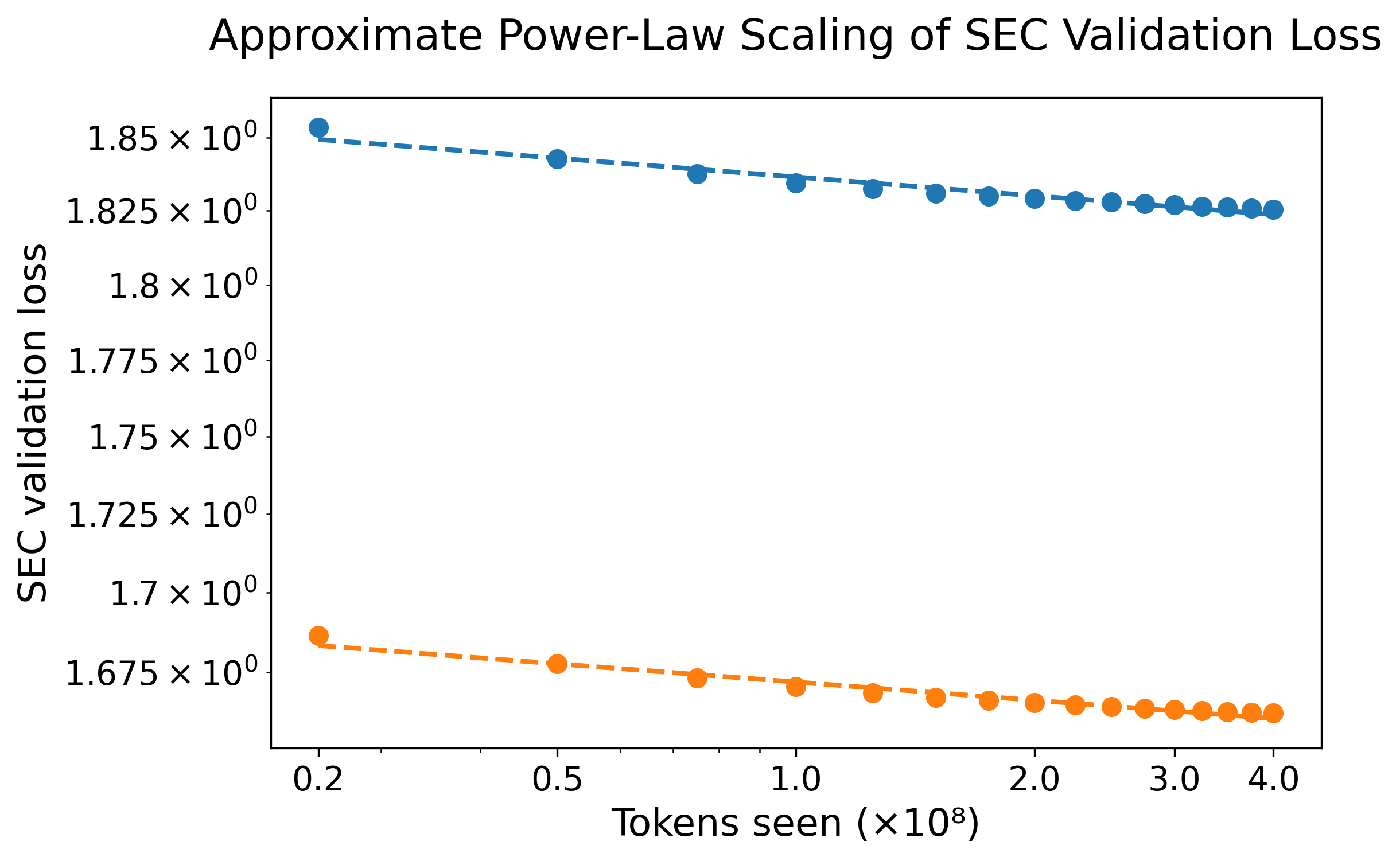}
    \vspace{-0.6em}
    \captionof{figure}{Power-law relationship between SEC validation loss and tokens seen}
    \label{fig:powerlaw}
\end{minipage}
\end{center}

\vspace{0.75em}

To assess whether continued pretraining follows predictable scaling dynamics, we fit a power-law model to validation losses recorded at all checkpoints, including the approximate 50M, 100M, 200M, and 400M token budgets. Although the raw loss curves appear nearly flat, plotting them in log-log coordinates reveals a clear linear trend for both model sizes.

The fitted exponents are shallow $(|b| \ll 1)$, characteristic of late stage continued pretraining rather than training from scratch. In this regime, the model already possesses broad linguistic structure, and additional domain-specific tokens primarily refine terminology and phrasing. The smooth log-log linearity indicates that SEC DAPT continues to yield real improvements, even when the gains are visually subtle in linear space.

Importantly, the shallow scaling rates reflect properties of the domain, not limitations of DAPT. SEC filings follow highly regular reporting conventions, with consistent structure and recurring terminology across years. Rather than reducing the value of the corpus, this regularity makes the domain efficient to learn, allowing models to acquire the essential financial patterns with comparatively modest token budgets. 

The fitted curves also allow coarse projections for larger foundation model scales. Extrapolating the 1B→3B trend to 32B or 70B models implies that achieving comparable relative improvements could require on the order of several billion domain tokens, still tractable compared to the trillion token budgets used for general purpose models. In other words, while larger models demand more data, the shallow exponent indicates that SEC language remains learnable and efficient even at foundation model scales.
Overall, the scaling law behavior demonstrates that SEC-specific DAPT follows a predictable improvement trajectory and that the underlying financial domain is well suited for continued pretraining, providing encouraging signals for future financial foundation model development.

\vspace{1em}

\begin{center}
\begin{minipage}{0.9\linewidth}
\centering

\captionof{table}{Estimated DAPT token requirements scaled from a 3B LLaMA baseline.}
\label{tab:dapt-scaling}

\begin{tabular}{lccc}
\toprule
\textbf{Model Size} & \textbf{Param Ratio vs 3B} & \textbf{Estimated Domain Tokens} \\
\midrule
3B  & 1×     & $4\times10^8$ (observed) \\
7B  & $\sim2.3\times$  & $\sim1$–$2\times10^9$ \\
13B & $\sim4.3\times$  & $\sim2$–$4\times10^9$ \\
32B & $\sim10.6\times$ & $\sim4$–$8\times10^9$ \\
70B & $\sim23\times$   & $\sim8$–$15\times10^9$ \\
\bottomrule
\end{tabular}

\end{minipage}
\end{center}

\vspace{1em}

\subsection{Mixed-Domain Validation Loss (General Domain Stability)}\label{sec:mixed-domain-validation-loss}

\vspace{0.6em}

\begin{center}
\begin{minipage}{0.9\linewidth}
    \centering
    \includegraphics[width=0.75\linewidth]{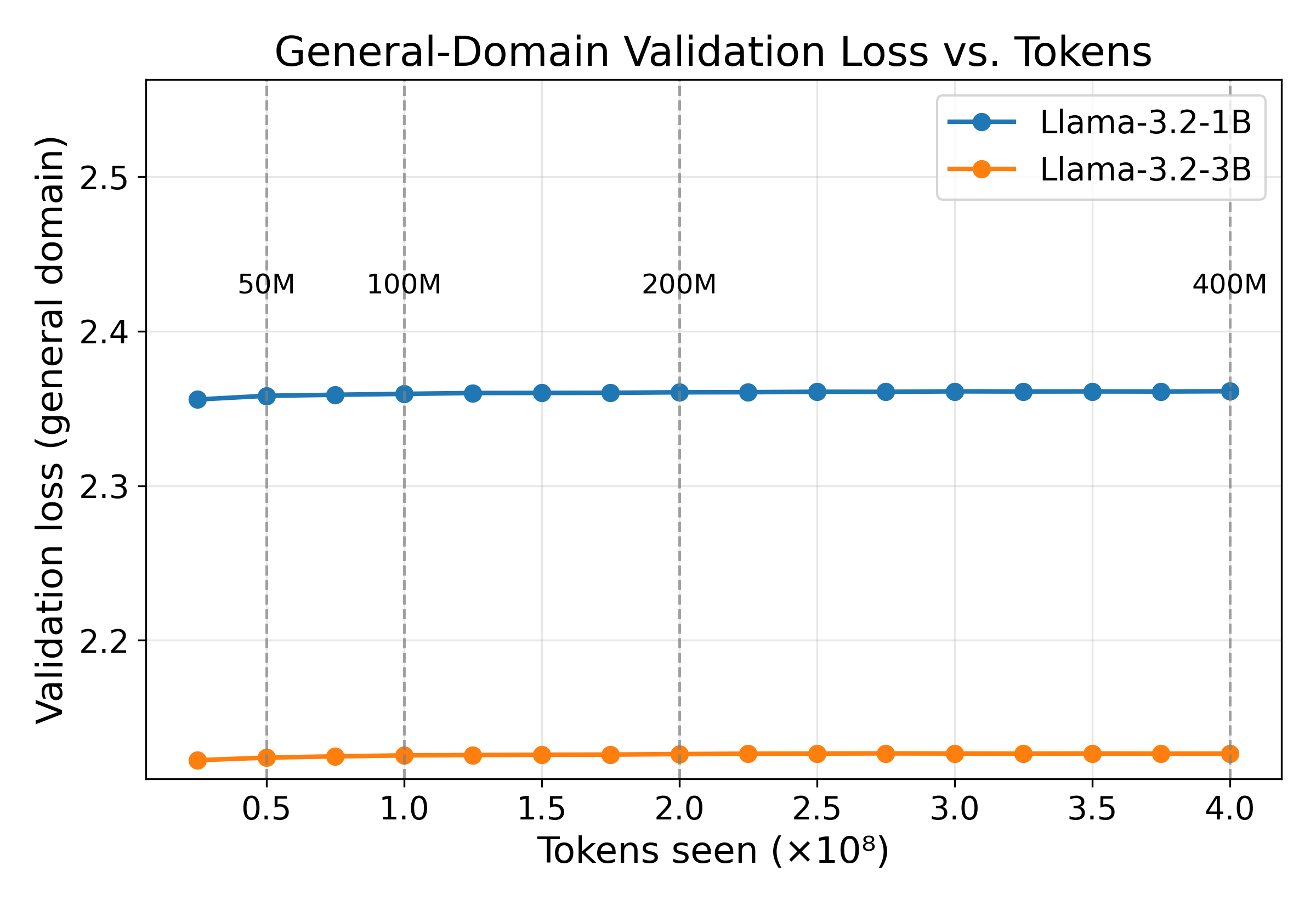}
    \vspace{-0.6em}
    \captionof{figure}{General-domain validation loss across continued pretraining}
    \label{fig:general-loss}
\end{minipage}
\end{center}

\vspace{0.75em}

Mixed-domain validation loss remains effectively constant across all token budgets for both the 1B and 3B models, indicating that continued pretraining on 400M tokens of SEC text does not degrade general domain performance. Although the 3B model achieves slightly lower loss overall, neither model exhibits any upward drift or trend suggestive of catastrophic forgetting.

The overall variation is extremely small, on the order of 0.01, well within expected noise for validation curves. This stability suggests that SEC language represents a relatively narrow domain shift. It introduces new financial terminology and structure but does not overwrite the broad linguistic representations learned during initial pretraining. As a result, models can specialize on financial text while preserving useful general domain capability.

\subsection{Data Efficiency Frontier}\label{sec:data-efficiency-frontier}

\vspace{0.6em}

\begin{center}
\begin{minipage}{0.9\linewidth}
    \centering
    \includegraphics[width=0.82\linewidth]{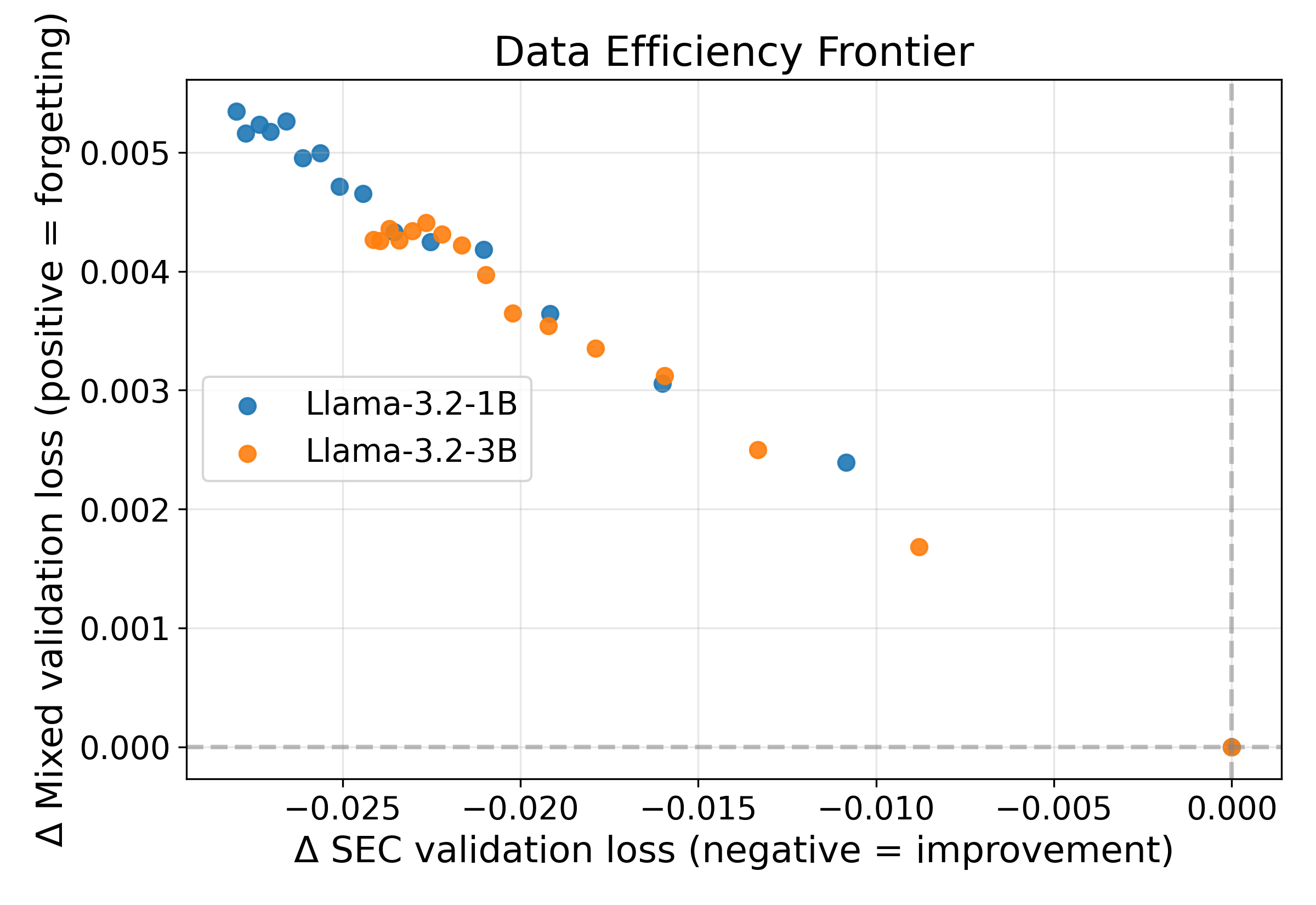}
    \vspace{-0.6em}
    \captionof{figure}{Data efficiency frontier comparing SEC- and general-domain loss changes}
    \label{fig:data-efficiency}
\end{minipage}
\end{center}

\vspace{0.75em}

The data efficiency frontier plots changes in SEC domain loss against changes in mixed-domain loss, providing a compact view of the trade off between specialization and forgetting. Both the 1B and 3B models move into the upper left region of the plane as the continued pretraining budget increases. SEC domain loss decreases, while general-domain loss remains effectively unchanged across all token budgets, with mixed-loss variation staying within the noise scale reported in Section 4.3.

The 3B model traces a slightly more favorable trajectory, tending to achieve larger SEC domain improvements for comparable levels of general-domain drift. Notably, 3B data points cluster within a narrow band just above a mixed validation loss of ~0.004, suggesting slightly less forgetting at similar rates of domain learning. This compact frontier highlights the domain efficiency of financial text and demonstrates that SEC-specific DAPT yields meaningful gains without compromising general purpose capability.

\section{Conclusion}\label{sec:conclusion}

Domain adaptive continued pretraining of Llama 3 models on SEC filings produces consistent but modest reductions in domain loss, with the 3B model showing slightly better data efficiency across all token budgets. Importantly, general domain performance remains stable, indicating minimal catastrophic forgetting even after 400M tokens of purely financial text.

Across both model sizes, validation losses follow a predictable log-log scaling trend, consistent with late stage continued pretraining rather than training from scratch. The resulting scaling exponents are shallow but coherent, reflecting the relatively low entropy, highly structured nature of SEC filings. This structure allows the models to learn the domain efficiently, though it also leads to diminishing marginal returns beyond roughly 300M domain tokens.

Overall, the results demonstrate that lightweight DAPT on financial filings is effective, enabling measurable specialization with comparatively modest data budgets. At the same time, achieving larger gains, particularly for substantially larger models, will likely require larger or more diverse financial corpora, or mixtures that incorporate adjacent domains such as earnings call transcripts, news, or regulatory commentary. These findings have direct implications for the development of future financial foundation models, pointing toward efficient specialization pathways that avoid trillion token pretraining requirements.

\section{Impact and Future Work}\label{sec:impact-and-future-work}

This study provides practical guidance for researchers and practitioners considering domain-adaptive continued pretraining on financial corpora. Several implications follow from our results:
\begin{itemize}
    \item \textbf{Token efficiency within the observed range, with implications for large-scale DAPT.}
    
    Within the 0-400M token regime, both models show clear improvement, with the majority of early gains occurring before ~250M tokens. This establishes the shape of the early scaling curve for financial continued pretraining and provides an empirical anchor point for designing larger DAPT efforts.
    \item \textbf{Need for downstream evaluation.}
    
    Future work should evaluate how loss reductions translate into task level performance on SEC QA, risk classification, structured extraction, or anomaly detection. This is essential for connecting scaling trends to practical financial use cases.
    \item \textbf{Scaling to larger models.}
    
    Extrapolations from the observed scaling behavior suggest that 7B-70B models could continue to benefit from SEC-specific DAPT, with domain token requirements still well below the budgets used for open domain foundation models.
    \item \textbf{Reducing drift through data mixtures.}
    
    Although catastrophic forgetting was minimal, mixed-domain or curriculum-based DAPT could further enhance stability for applications involving hybrid financial/general language.
\end{itemize}

Taken together, these findings contribute early evidence toward a scalable, data-efficient path for building financial foundation models, showing that meaningful specialization can be achieved without trillion token training regimes. The data efficiency frontier observed here provides a roadmap for constructing larger economic and regulatory reasoning models grounded in real world financial text.

\bibliographystyle{ieeetr} 
\bibliography{references}
\end{document}